\documentclass[10pt,conference,a4paper]{IEEEtran}

\usepackage{cite}
 \usepackage{graphicx}
\usepackage{amsmath}
\usepackage{amsfonts}
\usepackage{algorithm}
\usepackage{algorithmic}
\usepackage{array}
\usepackage{subfig}
\usepackage{color}
\usepackage{dsfont}
\usepackage{booktabs}
\usepackage{times}

\newcommand\given[1][]{\:#1\vert\:}
\begin{document}

\title{On Regularization Parameter Estimation \\under Covariate Shift}

\author{{\bf Wouter M. Kouw}\textsuperscript{1} \\ {\bf Marco Loog}\textsuperscript{1,2} \\
Pattern Recognition Lab\textsuperscript{1} \\
Delft University of Technology\\
Mekelweg 4, 2628CD Delft, the Netherlands\\
The Image Group\textsuperscript{2} \\
University of Copenhagen\\
Universitetsparken 5, DK-2100, Copenhagen , Denmark}

\maketitle

\begin{abstract}
This paper identifies a problem with the usual procedure for $L^2$-regularization parameter estimation in a domain adaptation setting. In such a setting, there are differences between the distributions generating the training data (source domain) and the test data (target domain). The usual cross-validation procedure requires validation data, which can not be obtained from the unlabeled target data. The problem is that if one decides to use source validation data, the regularization parameter is underestimated. One possible solution is to scale the source validation data through importance weighting, but we show that this correction is not sufficient. We conclude the paper with an empirical analysis of the effect of several importance weight estimators on the estimation of the regularization parameter.
\end{abstract}

\IEEEpeerreviewmaketitle

\section{Introduction}
In supervised learning, there is a (mostly implicit) assumption that the training data is an unbiased sampling of the underlying distribution of interest. However, that may not be the case. In a variety of problems there is often an unknown bias in the sampling procedure. These arise due to environmental effects, such as temperature in different genome sequencing centers \cite{shah2009ontology, mei2011gene, xu2011survey}, or due to the use of particular measuring instruments, such as types of cameras in computer vision \cite{saenko2010adapting, hoffman2013efficient}. This means the training dataset (source domain) and the test dataset (target domain) are technically generated by different distributions and generalization might no longer be possible. The challenge lies in using the labeled source data and the unlabeled target data to classify new target data; a problem setting often referred to as domain adaptation, transfer learning or sample selection bias \cite{cortes2008sample, quionero2009dataset, ben2010theory, margolis2011literature, moreno2012unifying}. Most research focuses on classifiers that incorporate information on the difference between the data in both domains, but unfortunately most of these approaches overlook the role of the regularization parameter.

Regularization is used to combat overfitting of complex models and is a vital component in most classifiers to ensure they generalize to unseen data. It consists of a trade-off between how well the classifier can discriminate training samples and how complex it must become to do so. This balance is described by the regularization parameter which is usually estimated by holding out a small subset of unseen labeled data and evaluating the trained classifier (cross-validation). However, since there are no labeled target samples available, it is not possible to construct a target validation set. If one were to alternatively construct a validation set from source data, the estimator converges in distribution to the source risk and not the target risk \cite{sugiyama2007covariate}. 

In this paper, we study how the generalization performance of a classifier behaves as a function of the regularization parameter and the domain dissimilarity. There are many factors that influence the value of the optimal regularization parameter, such as the moments of the class-conditional distributions in each domain (differences in variance, skewness, etc.), concept drift (different class priors in each domain), types of adapting classifiers (some require less regularization than others) and high-dimensional distribution estimation errors, but in this paper we focus on differences in variance between domains. The first correction that comes to mind consists of scaling the source validation risk with importance weights and although this remedies the problem somewhat, we show that the optimal regularization parameter for the target domain remains underestimated.

\subsection{Outline}
The paper is outlined as follows: section \ref{first} reviews the regularized empirical risk minimization framework and identifies the problem with selecting the optimal regularization parameter. Section \ref{second} illustrates the covariate shift setting and how the problem might be resolved there. Section \ref{fourth} considers several importance weight estimators with diverse properties while sections \ref{fifth} and \ref{sixth} present experimental evaluations of their estimates.

\section{Estimation Problem} \label{first}

\subsection{Preliminaries}
For a classification problem with a sample space $\Omega$ and an event space $\mathcal{F}$, the domains are biased samplings resulting in probability spaces with different probability measures $\mathbb{Q}$ and $\mathbb{P}$. Denote $\mathcal{X}$ as the random variable associated with the source domain, $(\Omega,\mathcal{F},\mathbb{Q})$, $\mathcal{Z}$ as the random variable associated with the target domain $(\Omega,\mathcal{F},\mathbb{P})$ and the classes as elements of $\mathcal{Y} = \{-1,+1\}$. Source data $X$ with labels $y$ consists of $n$ samples from $p_{\cal X, Y}$, denoted as $\{(x_{i},y_{i})\}_{i=1}^{n}$, and target data $Z$ with labels $u$ consists of $m$ samples from $p_{\cal Z, Y}$, denoted as $\{(z_{j},u_{j})\}_{j=1}^{m}$. A classifier is a function that takes as input data and outputs a class prediction, $h : \Omega \rightarrow \cal{Y}$.

\subsection{Regularized Risk Minimization}
The risk minimization framework allows one to construct classifiers through searching a class of hypothetical functions $H$ (e.g., linear) and selecting the one that minimizes the expected loss $\ell : \mathbb{R} \times \mathcal{Y} \rightarrow \mathbb{R}_{+}$. The source and target risk are defined respectively as:
\begin{align}
	R_{\mathcal{X}}(h) =& \int_{\Omega} \sum_{y \in \mathcal{Y}} \ \ell(h(x),y) \ p_{\cal X,Y}(x,y) \ \mathrm{d}x \label{Rx} \\
	R_{\mathcal{Z}}(h) =& \int_{\Omega} \sum_{y \in \mathcal{Y}} \ \ell(h(z),y) \ p_{\cal Z,Y}(z,y) \ \mathrm{d}z \label{Rz} \, .
\end{align}
Note that by virtue of the shared sample space $\Omega$ of the source and target domains, the differentials $x$ and $z$ are interchangeable, and that, for any $h$, they differ only through the joint probabilities $p_{\cal X,Y}$ and $p_{\cal Z,Y}$. The goal is to find a hypothesis $h$, based on a source sample, that will minimize the target risk.

Unfortunately, minimizing the empirical source risk with respect to $h$ often leads to a solution that does not generalize well to other samples (overfitting), let alone samples from another distribution. In order to restrict the classifiers ability to match the training sample as well as possible, a complexity term, in the form of the $L^p$-norm of the hypothesis, is added to the empirical risk during training:
\begin{align}
	\hat{R}_{\mathcal{X}}(h,X_T,\lambda) =&  \frac{1}{|X_T|} \sum_{t \in T} \ell(h(x_{t}), y_{t}) + \lambda \| h \|_{p}^{2} \label{rrX} 
\end{align}
where $T \subset \{1, \dots n\}$ denotes the set of indices used to select the training samples $\{(x_{t}, y_t)\} \subset \{(X, y)\}$, $|.|$ denotes the cardinality and $\| . \|_{p}$ denotes the $L^p$-norm. For the remainder of the paper, we will be working with the $L^2$-norm. 

The regularization parameter $\lambda$ trades off the average loss and the $L^2$-norm. It is usually estimated by defining a set of values $\Lambda$, training a classifier for each and selecting $\lambda \in \Lambda$ that minimizes the empirical risk evaluated on a disjoint validation dataset. The set of regularized classifiers can be denoted as:
\begin{align}
	h_{\Lambda} = \{h_{\lambda} = \underset{h \in H}{\arg \min} \ \hat{R}_{\mathcal{X}}(h,X_T,\lambda) \ | \ \lambda \in \Lambda \}  \label{hlam} \, .
\end{align}
where $h_{\lambda} \in h_{\Lambda}$ refers to the classifier that is trained using $\lambda \in \Lambda$. The regularization parameter space $\Lambda$ is often taken to be an exponentially increasing set of nonnegative values; for example $\{0, 0.01, 0.1, 1, 10, 100, 1000\}$. Note that regularization is added during training, but not during evaluation.

If we choose a quadratic loss function, $\ell(h(x),y) = (h(x) - y)^{2}$, with a linear hypothesis class, then the solution to minimizing equation \ref{rrX} with respect to $h$ is $h_{\lambda} = (X_{T}^{\top}X_{T} + \lambda I)^{-1}X_{T}^{\top}y_{T}$.

\subsection{Evaluation}
Evaluation of a classifier consists of its risk on a novel dataset. We will be studying two novel sets, the first being held out source data, because that is usually the only validation data available, and the second being target data, which is the actual measure of interest but is usually not available due to the lack of target labels. Taking the quadratic loss again, the resulting risk is also known as the Mean Squared Error. If we expand the square and plug in the held out source validation data $\{(X_{V},y_V)\}$, indexed by $V \subset \{1, \dots, n\}$ disjoint from the training set $V \cap T = \emptyset$, and the labeled target samples $\{(Z,u)\}$, the Mean Squared Errors are:
\begin{align}
	&\text{MSE}_V(h_{\lambda}) = \ 1 - \frac{2}{|X_V|} y_{V}^{\top}X_{V}h_{\lambda} + \frac{1}{|X_V|} h_{\lambda}^{\top}X_{V}^{\top}X_{V}h_{\lambda} \nonumber \\
	&\text{MSE}_Z(h_{\lambda}) = \ 1 - \frac{2}{|Z|} u^{\top}Zh_{\lambda} + \frac{1}{|Z|} h_{\lambda}^{\top}Z^{\top}Zh_{\lambda} \nonumber \, .
\end{align}
Cross-validation consists of holding out each source sample at least once, training a classifier on the remainder and evaluating on the held out validation set. One round of cross-validation is performed for each $h_\lambda \in h_\Lambda$ and the minimizer of the set with respect to the Mean Squared Error corresponds to the estimated regularization parameter.

\subsection{Problem}
For any $h$, the empirical source validation risk $\hat{R}_{\cal X}(h)$ converges to the true source risk $R_{\cal X}(h)$ by independently sampling validation sets infinitely many times:
\begin{align}
 	\mathbb{E}_{{X_V} \sim \mathcal{X}} \left[\hat{R}_{\cal X}(h,X_V)\right] = R_{\mathcal{X}}(h) \nonumber \, ,
\end{align}
which is unfortunately not equal to the true target risk $R_{\cal Z} (h)$. Furthermore, the larger the difference between $p_{\cal X,Y}$ and $p_{\cal Z,Y}$, the larger the difference between the minimizers of $R_{\cal X}(h_\lambda)$ and $R_{\cal Z}(h_\lambda)$ with respect to $\lambda$ and the larger the error in estimating the optimal regularization parameter. 

\section{Covariate Shift} \label{second}
A natural approach to designing a corrected cross-validation procedure, would be to employ some functional relation between the source and target risks. Fortunately, such a relation exists for a subset of the class of domain adaptation problems: if one makes the \emph{covariate shift} assumption that the class posterior distributions are equivalent in both domains, $p_{\cal Y \given Z} = p_{\cal Y \given X}$, then the target risk can be rewritten into a weighted source risk:
\begin{align}
	R_{\mathcal{W}}(h) =& \int_{\Omega} \sum_{y \in \mathcal{Y}} \ell(h(x),y) \ \frac{p_{\mathcal{Z}}(x)}{p_{\mathcal{X}}(x)} p_{\cal X,Y}(x,y) \ \mathrm{d}x \, . \nonumber
\end{align}
and the functional relation thus consists of weighting the source samples appropriately. It can be shown that under the additional assumption of a small domain discrepancy, this problem setting is learnable \cite{ben2010impossibility}. 

\subsection{Generating a covariate shift setting}
\begin{figure}[htb]
\centering
\includegraphics[width=.45\textwidth]{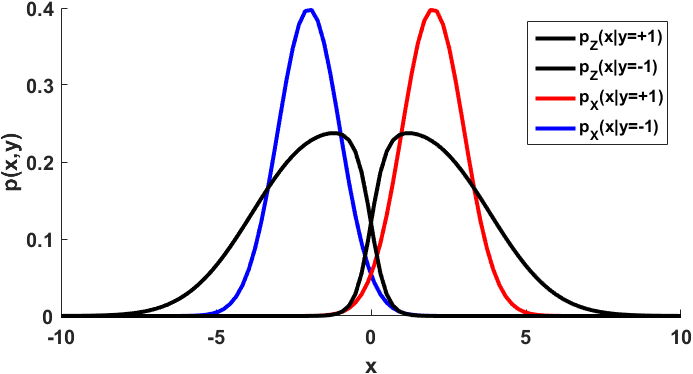} \\
\includegraphics[width=.45\textwidth]{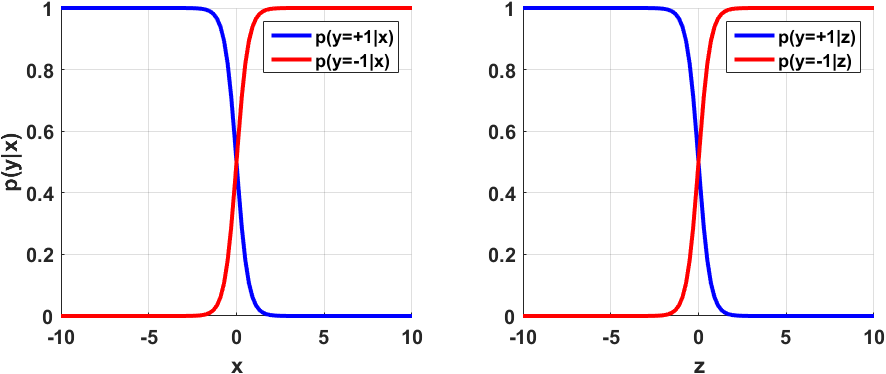}
\caption{An artificially generated 1-dimensional covariate shift problem. (Top) The class-conditional distributions in each domain. (Bottom) The class-posterior distributions of the source domain (left) and the target domain (right).}
\label{cv_problem}
\end{figure}

Since we are restricting the analysis to covariate shift settings, we need to generate such a problem. First, we choose a set of source class-conditional distributions $p_{\cal X \given Y}(x \given y)$, a set of priors $p_{\cal Y}(y)$ and compute the class posterior distributions $p_{\cal Y \given X}(y \given x)$ through Bayes' rule. Then, by choosing a different target distribution $p_{\cal Z}(z)$, multiplying by the derived class posterior distributions $p_{\cal Y \given Z}(y\given z)=p_{\cal Y \given X}(y \given x)$ and inverting the Bayes' rule, the class-conditional target distributions $p_{\cal Z \given Y}(z \given y)$ are obtained. Figure \ref{cv_problem} (top) visualizes an example of this problem for Gaussian class-conditional distributions. We plotted the labeled source distributions in red and blue with the unlabeled target distributions in black. The class posteriors of this problem are plotted in Figure \ref{cv_problem} (bottom), and are equivalent. An artificial dataset can be generated by sampling from these distributions, either through inverse transform sampling or rejection sampling.

\begin{figure}[htb]
\centering
\includegraphics[width=.45\textwidth]{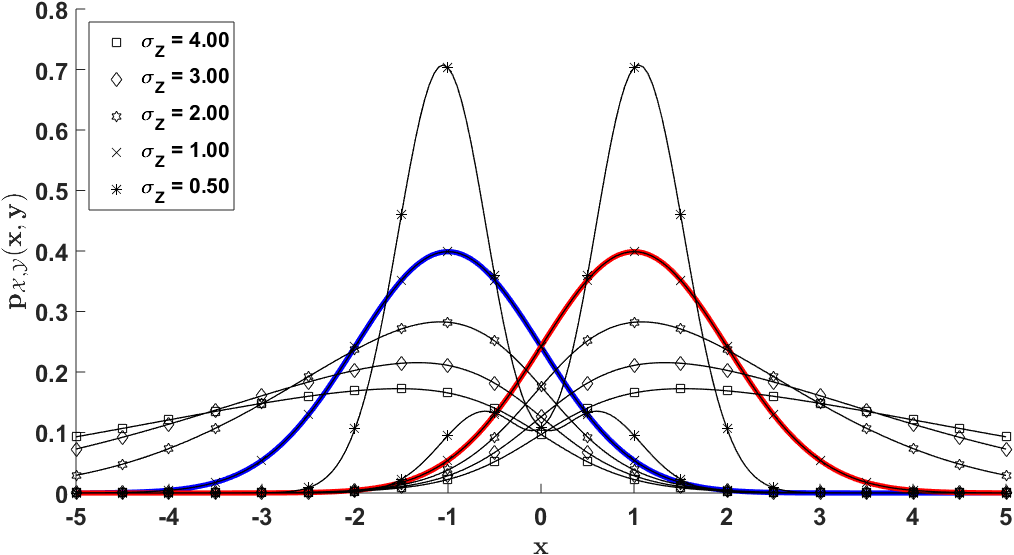} \\
\includegraphics[width=.45\textwidth]{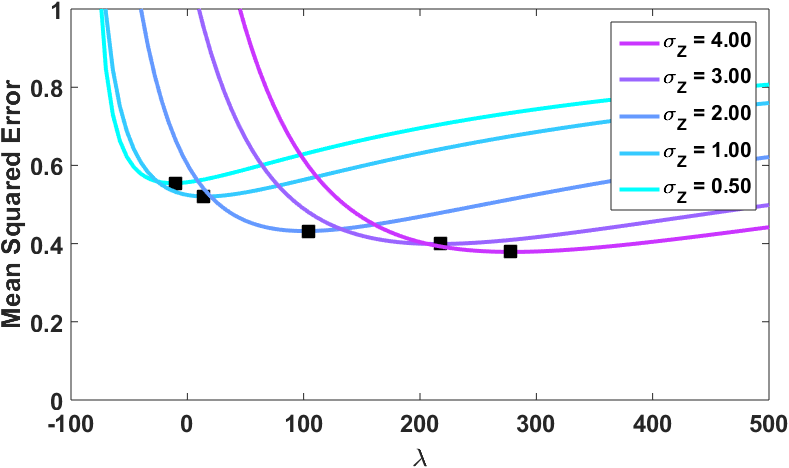} 
\caption{(Top) 5 covariate shift problems, with the target variance $\sigma^2_Z \in \{0.5, 1,  2, 3, 4\}$. (Bottom) The corresponding target MSE curves. The black squares denote the minima of these curves.}
\label{cvshift_scale_mse}
\end{figure}

If we fix the source class-conditional distributions to be Gaussian distributions, with the blue class as $\mathcal{N}(-1,1)$ and the red class as $\mathcal{N}(1,1)$, then we can generate 5 problem settings by choosing 5 different target distributions. Figure \ref{cvshift_scale_mse} (top) shows 5 Gaussian target distributions with equal means but with different variances $\sigma^2_{\cal Z} \in \{0.5, 1,  2, 3, 4\}$. If we train a classifier based on the source class-conditional distributions and evaluate it using the target MSE, then it becomes apparent that the difference between the minimizer of the source risk and the target risk starts to increase as the difference between the distributions start to increase. Figure \ref{cvshift_scale_mse} (bottom) plots the MSE as a function of $h_{\Lambda}$ for the 5 covariate shift problems, with the minimum for each marked with a black square. Note that for $\sigma_{\cal Z}^2 = 1$ the distributions are equivalent and its minimizer is equivalent to the minimizer of the source risk. The curves show a gradual increase in the minimizers as the variance increases.

\subsection{Difference in MSE curves}
If we minimize the MSE curves of both the source validation risk and the target risk with respect to the trained regularized classifier $h_\lambda$, we obtain:
\begin{align}
	h_{\hat{\lambda}_V} =& \ \ (X_{V}^{\top}X_{V})^{-1}X_{V}^{\top}y_{V} \nonumber \\
	h_{\hat{\lambda}_Z} =& \ \ (Z^{\top}Z)^{-1}Z^{\top}u \nonumber
\end{align}
where the subscripts $\hat{\lambda}_V$ and $\hat{\lambda}_Z$ denote the optimal regularization parameter according to the source validation and the target risk, respectively. Studying these two forms, we see that these estimates of $\lambda$ differ mainly through their data inner products (i.e., the uncentered, unnormalized covariance matrices). To illustrate this point, we can decompose the data through a singular value decomposition, allowing us to express the minimizers as:
\begin{align}
	h_{\hat{\lambda}_V} =& \ \ V_{V}D_{V}U_{V}^{\top}y_{V} \nonumber \\
	h_{\hat{\lambda}_Z} =& \ \ V_{Z}D_{Z}U_{Z}^{\top}u  \nonumber
\end{align}
where the diagonal matrices $D_{V}$ and $D_{Z}$ consist of the normalized singular values $D_{V,ii} = \alpha_{V,i} / \alpha^{2}_{V,i}$ and $D_{Z,ii} = \alpha_{Z,i} / \alpha^{2}_{Z,i}$. Apart from a change of basis from $V_V$ to $V_Z$ and $U_V$ to $U_Z$, the difference lies in the scale of the eigenvalues.

If we were to apply a scaling operation to the validation risk, then the difference between these curves can be minimized. Finding the optimal regularization parameter for the target domain will then be equivalent to finding the optimal regularization parameter for the scaled validation risk.

\subsection{Importance Weighted Validation}
Sugiyama et al. (2007) employ just such a scaling transformation in the form of importance weighting the validation risk, with the weights as estimates of the ratio of data marginals $p_{\cal Z}(x) / p_{\cal X}(x)$ \cite{sugiyama2007covariate}. These weights scale the risk of each individual validation sample separately. This leads to an importance weighted source validation risk as follows:
\begin{align}
	\text{MSE}_{W}(h_{\lambda}) =& \ 1 - \frac{2}{|X_V|} y_{V}^{\top}WX_{V}h_{\lambda} + \frac{1}{|X_V|} h_{\lambda}^{\top}X_{V}^{\top}WX_{V}h_{\lambda} \nonumber \, 
\end{align}
where $W$ is a matrix with the importance weights $w(x_{v})$ as its diagonals. This formulation has the following minimizer:
\begin{align}
	h_{\hat{\lambda}_W} =& \ \ (X_{V}^{\top}WX_{V})^{-1}X_{V}^{\top}Wy_{V} \nonumber \, .
\end{align}

This ratio of probabilities can have a very large variance, depending on how likely it is to evaluate it for either extremely large target probabilities or extremely small source probabilities. Furthermore, in the small sample size setting, estimation errors increase the likelihood of encountering a numerical explosion, such as when 10 samples are drawn that lie so close together that the estimated target distribution resembles a Dirac distribution. Lastly, the cross-validation estimator has its own variance \cite{markatou2005analysis} which is now directly affected by the variance of the importance weight estimator. For a better understanding of the behavior of an importance weighted cross-validation estimator, we performed a number of experiments with a large diversity of weight estimators in the following section.

\section{Experiments}
We conducted an experiment on an artificial problem setting and one on a typical real-world domain adaption problem where there is no knowledge on whether the covariate shift assumption holds. Our goal is to evaluate the ability of a number of both parametric and nonparametric importance weight estimators to correctly estimate the optimal regularization parameter in the target domain. These experiments illustrate that a large diversity of existing estimators tends to underestimate the optimal target parameter.

\subsection{Importance weight estimators} \label{fourth}
We selected four importance weight estimators with a diverse set of behaviors.

\subsubsection{Ratio of Gaussians ({\sc rG})} A baseline method of estimating the marginal data ratio through modeling each sample set with a separate Gaussian distribution \cite{shimodaira2000improving}: 
\begin{align}
	w_{\text{{\sc rG}}} = \frac{\mathcal{N}(x \given \hat{\mu}_{T}, \hat{\sigma}^{2}_{T})}{\mathcal{N}(x \given \hat{\mu}_{S}, \hat{\sigma}^{2}_{S})} \nonumber \, ,
\end{align}
where $\mathcal{N}$ denotes the Gaussian distribution function, the $\hat{\mu}$'s denote the estimated means of the subscripted sample set and the $\hat{\sigma^2}$'s denote the estimated variance of the subscripted sample sets. Note that the data marginals in our problem are actually Gaussian and that this is thus the correctly specified model. 

\subsubsection{Kullback-Leibler Importance Estimation Procedure ({\sc kliep})} This popular method is based on minimizing the Kullback-Leibler divergence between the reweighted source samples and the target samples \cite{sugiyama2008direct}:
\begin{align}
	w_{\text{{\sc KLIEP}}}  = \underset{w \in W}{\arg \max}& \ \sum_{j=1}^{m} \log \ \sum_{i}^{n} w_i K(x_{i}, z_{j}) \nonumber \, , \\
		\text{s.t.}& \ \sum_{i}^{n} w_i K(x_{i}, z_{j}) = n \nonumber
\end{align}
where the constraint avoids numerical explosions. For $K$ we chose a Gaussian kernel, with the kernel width $\sigma$ estimated through a separate 3-fold cross-validation \cite{sugiyama2008direct}.

\subsubsection{Kernel Mean Matching ({\sc kmm})} Another popular weight estimator that is motivated by assigning weights that minimize the Maximum Mean Discrepancy (MMD) between the reweighted source and the target samples \cite{huang2006correcting}. The MMD is the distance between the means of two sets of samples under a worst-case transformation (one that pushes them as far away as possible):
\begin{align}
	w_{\text{{\sc KMM}}} = \ \underset{w \in W}{\arg \min}& \ \frac{1}{2} w^{\top}K(x,x')w - \frac{n}{m}\sum_{j=1}^{m}K(x,z_{j})w \nonumber \, , \\
	\text{s.t.}& \quad w_i \in [0 \ B] \nonumber \\
	& \quad |\sum_{i=1}^{n} w_i - n | \leq n \epsilon \nonumber
\end{align}
where the constraints ensure that the weights are non-negative, bounded above and roughly sum to the sample set size. For the kernel, we selected a radial basis function with Silverman's rule of thumb for bandwidth selection. Huang et al. recommend setting epsilon to $B / \sqrt{n}$ ensuring that the allowed deviation from the sample size depends on both the upper bound for each weight and the sample set size itself.

\subsubsection{Nearest Neighbour ({\sc nn})} Lastly, we have a nonparametric estimator based on a Voronoi tessellation of the space \cite{loog2012nearest}. The procedure consists of assigning a weight to each source sample based on the number of target samples that are nearest neighbors of it and is proportional, up to the ratio of sample sizes, to the ratio of marginal distributions. It is expressed as:
\begin{align}
	w_{\text{NN}} = |C_{i} \cap \{z_{j}\}_{j=1}^m| + 1 \nonumber \, ,
\end{align}
where $C_{i}$ refers to the Voronoi cell of sample $x_{i}$. The tessellation can be smoothed by adding a value of 1 to each cell, a technique also known as Laplace smoothing.

\subsection{Artificial data} \label{fifth}
Our first experiment consists of an evaluation of different importance weight estimators and their resulting minimizers of $h_{\Lambda}$. The set $h_{\Lambda}$ was constructed through linear least squares classifiers $h_{\lambda} = (X_{T}^{\top}X_{T} + \lambda I)^{-1}X_{T}^{\top}y_{T}$, with a range for $\Lambda$ from -100 to 500. For the source data, we drew 100 samples from two Gaussian class-conditional distribution with means $\mu_{\cal X} \in \{-1, 1\}$ and unit variances $\sigma_{\cal X}^{2} = 1$. The target class-conditional distributions have the same mean $\mu_{\cal Z}  \in \{-1, 1\}$, but with a different set of variances $\sigma_{\cal Z}^{2} \in \{0.1, 0.5, 1, 2, 3, 4\}$. The ratio of the marginal distributions is sensitive in regions of low probability of the source distribution; really small probabilities in the denominator explode the weight value. Therefore, we expect the minimizers of the importance weight estimators to be close to the target minimizer for smaller target variances $\sigma_{\cal Z}^{2} < \sigma_{\cal X}^{2}$. Consequently, we expect erratic behavior for target variance larger than the source variance $\sigma_{\cal Z}^{2} > \sigma_{\cal X}^{2}$. Table \ref{diff_Ll} displays the minimizers of $h_{\Lambda}$ for the source validation risk, for the different importance weight estimators, for the actual ratio of marginals $p_{\cal Z}(x) / p_{\cal X}(x)$, and for the empirical target risk. They are the means and standard errors over 100 repeats.

\begin{table}[th]                                                               
\caption{The mean and standard error of the estimated regularization parameter $\hat{\lambda}$ for different importance weight estimators and an increasingly larger target variance in a covariate shift problem.}    
\label{diff_Ll} 
\setlength{\tabcolsep}{3pt}
\setlength\extrarowheight{3pt}
\begin{center}
\begin{tabular}{l |r r r r r r}  
$\sigma_{\cal Z}^{2}$ & 0.1 & 0.5 & 1.0 & 2.0 & 3.0 & 4.0 \\
 \midrule
$h_{\hat{\lambda}_V}$ & 4 (19) & 3 (20) & 5 (20) & 3 (20) & 4 (20)  & 4 (19) \\
\midrule
$\hat{w}_{\text{{\sc rG}}}$ 	& -15 (20) & -10 (19) & 6 (17) & 30 (24) & 55 (41) & 58 (36)  \\
$\hat{w}_{\text{{\sc KLIEP}}}$ 	& -23 (33) & -2 (20) & 3 (20) & 0 (17) & 4 (24) & 17 (25)  \\
$\hat{w}_{\text{{\sc KMM}}}$ 	& 22 (24) & 17 (26) & 11 (25) & -1 (24) & -15 (21) & -14 (19) \\
$\hat{w}_{\text{{\sc NN}}}$ 	& -18 (28) & -13 (21) & 4 (23) & 33 (24) & 53 (24) & 64 (26)  \\
\midrule
$p_{\cal Z} / p_{\cal X}$ & -46 (47) & -33 (21) & 3 (18) & 46 (44) & 72 (50) & 77 (50)  \\
\midrule 
$h_{\hat{\lambda}_Z}$ & 179 (137) & -24 (21) & 5 (21) & 102 (28) & 207 (38) & 296 (45)
\end{tabular}  
\end{center}   
\end{table}

It seems that all importance weight estimators as well as the true ratio of marginals underestimate the target risk minimizer. Furthermore, it seems that $\hat{w}_{\text{{\sc kmm}}}$ leads to increasingly smaller minimizers for an increasing target variance. Even though $\hat{w}_{\text{{\sc KLIEP}}}$ is increasing, it still underestimates the true value the most. {\sc $\hat{w}_{\text{{\sc rG}}}$} is the most accurate one, but that will probably not be the case if the marginal distributions are not Gaussian anymore (i.e., model misspecification). {\sc $\hat{w}_{\text{{\sc NN}}}$} is the other most accurate one and lies closest to true importance weights. Considering that it does not rely on an assumption of normality, it might be the preferred estimator in a more general setting.

\subsection{Heart disease} \label{sixth}
The artificial data represents a case where we know exactly what the dissimilarity is between domains and whether assumptions are valid. However, it is also interesting to evaluate on data where we do not have this knowledge. For this we have selected a UCI dataset \cite{Lichman:2013} on medical data where the domain dissimilarity is caused by a geographically biased sampling of patients. The goal is to classify the presence of a heart disease based on symptoms. The four domains correspond to hospitals in `Cleveland', `Virginia', `Hungary' and `Switzerland', containing 303, 200, 294 and 123 samples each respectively. There are a total of 14 symptoms, but 2 contained so much missing data ($> 99\%$) that these were removed from the set. All other missing data was imputed with $0$ values after z-scoring, i.e. subtracting the mean of each feature and normalizing by its standard deviation. Table \ref{hdis_Ll} displays the minimizers found by the importance weight estimators compared with those found by the unweighted source validation risk  $h_{\hat{\lambda}_V}$ and the empirical target risk $h_{\hat{\lambda}_Z}$, for all combinations of treating one hospital as the source domain and another as the target. Shown are the means and standard errors over 10 repetitions.

\begin{table}[th]                                                               
\caption{Heart disease dataset. Mean and standard error of the estimated regularization parameter $\hat{\lambda}$ for different importance weight estimators. The letters are abbreviations of the 4 hospitals: C='Cleveland', V=`Virigina', H='Hungary' and S='Switzerland.}    
\label{hdis_Ll} 
\setlength{\tabcolsep}{5pt}
\setlength\extrarowheight{0pt}
\centering                                               
\begin{tabular}{l l | r r r r r| r}   
{\sc X} & {\sc Z} & $h_{\hat{\lambda}_V}$ & $\hat{w}_{\text{{\sc rG}}}$ & $\hat{w}_{\text{{\sc KLIEP}}}$ & $\hat{w}_{\text{{\sc KMM}}}$ & $\hat{w}_{\text{{\sc NN}}}$  & $h_{\hat{\lambda}_Z}$\\
\midrule
C & V & 1 (5) & -1 (8) & 1 (5) & 9 (13) & 2 (5) & 500 (0) \\
C & H & 1 (4) & 4 (6) & 1 (6) & 2 (14) & 4 (5) & 500 (0) \\
C & S & 4 (6) & 7 (9) & 0 (5) & 9 (12) & -1 (9) & 500 (0) \\
V & H & 5 (5) & 9 (13) & 3 (6) & 2 (5) & 7 (9) & 417 (66) \\
V & S & 3 (4) & -1 (12) & 3 (6) & 2 (3) & 7 (8) & -60 (284) \\
H & S & 3 (6) & 34 (48) & 3 (8) & 44 (40) & 4 (6) & 500 (0)\\
V & C & 4 (5) & -1 (9) & 2 (4) & 2 (3) & 4 (4) & 500 (0) \\
H & C & 1 (5) & 0 (7) & 2 (7) & 31 (29) & 0 (6)  & 500 (0) \\
S & C & 2 (4) & -1 (4) & 2 (4) & 1 (3) & 2 (4) & 488 (30) \\
H & V & 4 (4) & -15 (14) & 4 (7) & 25 (43) & 4 (9) & 500 (0) \\
S & V & 2 (4) & -1 (4) & 1 (7) & 1 (3) & 4 (4) & -95 (253) \\
S & H & 0 (4) & 4 (8) & 2 (6) & 0 (5) & 4 (4) & 289 (89)
\end{tabular}                                                                   
\end{table}

The results show that also for real datasets all importance weight estimators underestimate the optimal target regularization parameter. Note that the standard errors are 0 for all $h_{\hat{\lambda}_Z}$ that have value 500, which is because 500 is the right boundary of the set $\Lambda$. Extending the range further would produce even larger values for the optimal target regularization parameter. It seems that $\hat{w}_{\text{{\sc KMM}}}$ is the best performing estimator here. $\hat{w}_{\text{{\sc rG}}}$ also produces reasonable results, but that would probably not be the case if we had not z-scored each feature first. That ensures an overlap of the regions with high probability mass in each domain. The other estimators seem to find weight values close to 1, as they are not very different from the unweighted source validation risk.

\section{Discussion}
Considering the significance of regularization to generalization, it would be interesting to further study factors that influence the difference between the risk minimizers in each domain. At the moment we assume that no concept drift has occurred (a difference between class priors in each domain), but if this assumption is violated then the difference in scale depends on the two dominant classes in each domain. The minimizers of the MSE would be dominated by the proportions of samples that belong to one class, which can get very complicated in the multi-class setting. Furthermore, it would be interesting to describe the minimizers in terms of general measures of domain dissimilarity, such as the discrepancy distance \cite{mansour2009domain} or the ${\cal H}$-divergence \cite{ben2010theory}.

The main difficulty in estimating the appropriate weights lies in the fact that it is hard to estimate exactly how the two domains differ from each other. Most adaptation approaches are sensitive to only a particular type of relation between domains or rely on assumptions that can not be checked in advance. Furthermore, estimation errors tend to propagate. For instance, if the distributions of each domain's data marginals are poorly estimated, then the importance weights explode, leading to a more erroneous estimate of the optimal target regularization parameter. In domain adaptation settings with so many sources of uncertainty, it seems that simple methods work best.

\section{Conclusion}
We have shown an empirical analysis of regularization parameter estimation in the context of differing variances in covariate shift problems. It seems that the generalization performance of an unadapted source classifier can be improved by importance weighting the source validation risk. However, most popular weight estimators underestimate the optimal target regularization parameter.

\section*{Acknowledgment}
This work was supported by the Netherlands Organization for Scientific Research (NWO; grant 612.001.301).

\bibliographystyle{IEEEtran}
\bibliography{kouw_icpr16a}

\end{document}